# Digital diglossia: Arabic between X and Facebook

Fahad Al Hussen, King Saud University, Riyadh, Saudi Arabia
Mohammed Q. Shormani, Ibb University, Ibb, Yemen
shormani@ibbuniv.edu.ye/https://orcid.org/0000-0002-0138-4793

**Abstract**

This study highlights the distribution of Standard Arabic (SA; H(igh) variety) and Colloquial Arabic (CA; L(ow) variety) across X and Facebook. 16754 public posts were collected via Python, with 10000 retained as the net dataset. Posts were classified into 7 discourse categories: *politics, technology, science, business, culture, fun,* and *sports*. Bivariate analyses, including Chi-square tests and Cramér's V (CV), examined associations among platform, discourse category, and diglossic choice, while binary logistic regression with Platform × Discourse Category interactions tested whether these associations varied across platforms. Findings reveal that there are significant associations between discourse category and diglossic choice on X, $\chi^2(6, N = 5000) = 600.35$, $p < .001$, $CV = .347$, and Facebook, $\chi^2(6, N = 5000) = 1249.52$, $p < .001$, $CV = .500$. Across platforms, platform was also associated with diglossic choice, $\chi^2(1, N = 10000) = 262.16$, $p < .001$, $CV = .162$. Binary logistic regression further shows higher odds of SA use on X than Facebook in the political reference category ($OR = 1.31$, $p = .0028$), with significant platform-by-domain interactions for Culture ($OR = 2.65$), Fun ($OR = 6.34$), Sports ($OR = 26.71$), Science ($OR = 0.41$), and Technology ($OR = 0.71$). The study concludes that the diglossic use of SA and CA contributes to the growing body of research on digital discourse, unveiling that the digital age reshapes but does not erode diglossic boundaries, giving rise instead to a reconfigured *digital diglossia*.

## 1. Introduction

Diglossia is a linguistic phenomenon whereby two (or more) varieties are used in communication (Ferguson 1959). Ferguson refers to a diglossic context as having both function and stability, the former refers to the strict, complementary division where H(igh) varieties are used exclusively for formal, written, and institutional domains, while L(ow) varieties are reserved for informal, everyday communication. Stability, however, describes the long-term equilibrium of this arrangement, asserting that this rigid functional split persists comfortably over centuries without the two varieties merging or displacing one another (see also Saiegh-Haddad and Henkin-Roitfarb 2014; Saiegh-Haddad and Spolsky 2014).

Arabic is an ideal example of this phenomenon where a diglossic situation manifests clearly: 2 distinct varieties, viz., H and L of the language coexist within the Arab community, each serving different social functions. Recently, these ideas have been enriched by a number of scholars (see e.g., Taha Thomure et al. 2025; Khamis-Dakwar et al. 2019; Khamis-Dakwar et al. 2022; Froud and Khamis-Dakwar 2021). However, with the advent of the digital age (DA), coincided with vast and fast development in technology, internet, and artificial intelligence, several questions emerge as to: What changes has DA brought about to Arabic diglossia? (cf. Yuldasheva and Sidikova 2025), What type of discourse has DA brought about? (Thurlow and Mroczek 2011), What is the diglossic discourse in this age? (Esposito and KhosraviNik 2023), Does Critical Discourse Analysis (CDA) approach fit to analyze digital diglossic discourse? (Vásquez 2022) and, above all, can we employ corpus-based approach to analyzing this phenomenon? (Shormani and Alenezi 2026). To the best of our knowledge, perhaps no research has tackled this topic addressing such questions, and hence this study serves to fill this gap.

Digital diglossia refers to the continuation and transformation of traditional diglossic patterns within online and social media contexts (Alkhamees et al. 2019; Bassiouney 2009). In the Arabic-speaking world, diglossia often manifests as the use of Standard Arabic (SA) in formal settings and colloquial Arabic (CA) dialects in everyday communication (cf. Thurlow and Mroczek 2011). SA is the official, formal, pan-Arabic identity variety, used in press, TV, paperwork, State dealings (Shormani 2023). CA, however, is used in everyday dealings, homes, markets, life affairs, business advertisement, and entertainment. CA comprises all regional dialects (every Arab State has its own dialect), and within each State there exist several subvarieties. For example, Saudi Arabic is situated within the Gulf dialect; however, it has its own linguistic features which make it different from other Gulf dialects such as Kuwaiti Arabic Omani Arabic. And within Saud Arabic, there are several subvarieties including Najdi Arabic, Taifi Arabic, Hijazi Arabic, etc. (see, e.g., Aljasir 2020; Omar and Ilyas 2018)

However, perhaps until now, it is difficult to determine which variety of Arabic dominates social media platforms (SMPs). Social media has become a central arena for communication, politics, identity expression, and ideological negotiation across the world, in general, and the Arab world, in particular (Shirazi 2013; Sarkhoh and KhosraviNik 2020). Digital platforms such as X, formerly Twitter, and Facebook serve not only as channels of information exchange but also as dynamic spaces where Arabic speakers engage in diverse forms of discourse. Nowadays, social media allows for immediacy, interactivity, and multimodality, shaping how individuals present themselves, construct arguments, and participate in collective discussions (Androutsopoulos 2014; Herring 2019). For Arabic in particular, social media interaction raises important sociolinguistic and discursive questions. Given the diglossic nature of Arabic (Ferguson 1959; Badawi 1973), online communication often blends SA, CA, and code-switching with other languages such as English and French (Warschauer et al. 2002). This hybrid usage reflects both linguistic flexibility and social positioning, as speakers strategically shift codes depending on communicative goals, audience culture, education, societal roles, and platform constraints. Social media thus provides a unique window into the dynamics of Arabic diglossia, identity construction, and discourse in the digital arena.

Although there is a growing body of research on Arabic digital discourse (ADD), existing studies tend to focus on single platforms or on broad issues of language choice. For instance, studies of Arabic on X have highlighted its role in political mobilization, identity construction, and hashtag activism (Mubarak et al. 2017, cf. also Zappavigna 2012). Research on Facebook, however, has emphasized community-building, personal expression, and cultural discourse. Comparative studies across platforms remain rare, particularly in the Arabic context, where platform-specific constraints, such as brevity on X vs. narrative affordances on Facebook, can shape discourse in distinctive ways (Nwagbara 2025).

With this in mind, there seems to be a need to understand how Arabic diglossia operates within contemporary digital communication, where written interaction increasingly accommodates linguistic practices traditionally associated with both formal and informal contexts. This study thus bridges a critical gap by examining how Arabic discourse practices diverge between and within X and Facebook, focusing on linguistic variety, discourse functions, and thematic and digital orientations. We employ a corpus-based approach to studying this phenomenon, as it provides a systematic method for investigating these patterns. We utilize Python scraping libraries to extract data from Facebook and X, and perform Cohen's kappa and Chi-square tests. 2 coders were recruited to annotate/label the net data, and agreement was at $\kappa = 0.79$. We divided the dataset of tweets/posts into 7 categories including *politics, technology, science, business, culture, fun,* and *sports,* and analyze both datasets in terms of SA and CA. In this work, SA and CA are identified through a number of criteria including orthography, lexicon, morphology and syntax, and discourse markers. SA is commonly characterized by standardization of these features, while CA is identified through dialectal lexicon, non-standard linguistic features, and dialect-specific discourse markers. Our research work thus contributes to a more context-sensitive understanding of digital environments which do not simply eliminate Arabic diglossia but reconfigure the relationship among register, discourse domain and platform. Thus, it endeavors to provide plausible answers to the following questions:

1. Which Arabic variety is dominant on X and Facebook, SA or CA?
2. How is Arabic diglossia involved in digital discourse?
3. What are the most discursive themes expressed in diglossic digital discourse?

Thus, the remainder of this article is set up as follows. In section 2, we review the relevant literature on Arabic diglossia, social media discourse, and corpus-based approaches to linguistic analysis, providing the theoretical and empirical context. In section 3, we present the methodology, including corpus construction, data collection from X and Facebook, and approach of analysis for examining the distribution and features of H and L Arabic varieties. In section 4, we outline and tabulate the results, highlighting patterns of language use, and platform-specific trends in Arabic diglossia. Section 5 discusses the study findings quantitatively and qualitatively, and recommendations for future research. In section 6, we conclude the paper, providing implications for understanding diglossia in ADD.

## 2. Theoretical framework and literature review

### 2.1. Arabic digital diglossia

The notion *diglossia* was first introduced by Ferguson (1959), who described it as a sociolinguistic situation where 2 distinct varieties of the same language coexist within one speech community, each assigned to specific social functions. Arabic situation is an optimal context, manifesting in SA as the H variety, and CA as the L variety (or varieties), thereby maintaining the coexistence of SA and CA regional dialects (Khamis-Dakwar et al. 2019; Khamis-Dakwar et al. 2022). Diglossia remains central to any research on Arabic discourse (Eisele 2021; Leikin et al. 2014), the H variety is used in formal and institutional settings such as education, religion, administration, and literature, while the L variety is reserved for everyday conversation, family interactions, and informal domains (see e.g., Khamis-Dakwar et al. 2019; Khamis-Dakwar et al. 2022). According to Ferguson, there are four classical diglossic languages, viz., Arabic, Modern Greek, Swiss German, and Haitian Creole. He argues that diglossia is marked by a stable functional distribution, differences in acquisition (with the L variety learned natively and the H variety through schooling), and an asymmetry in prestige, with the H variety considered superior. Fishman (1967) broadened the definition to include cases where the H and L roles are played by different languages rather than 2 varieties of the same language, as in many multilingual societies. Subsequent research has further refined the concept, viewing diglossia not as a rigid binary but as a continuum of linguistic practices shaped by social, cultural, and ideological forces (Eisele 2013).

Arabic diglossia is considered the most prototypical and enduring example of Ferguson's (1959) model. The H variety corresponds to SA (and classical Arabic in Ferguson's sense) used in formal education, media, literature, religion, and official communication. The L varieties are represented by the numerous spoken dialects such as Saudi Arabic, Levantine, Egyptian, and Yemeni Arabic which serve as the native languages of Arabs across different regions/countries (cf. Shormani 2019). Unlike SA, these dialects are not standardized or formally taught. They are acquired naturally in childhood (Shormani 2023). The diglossic situation in Arabic is particularly complex due to the wide structural gap between SA and the dialects. Some scholars (Ferguson 1959; Badawi 1973; Eisele 2013) describe it as a "diglossic continuum," where multiple intermediate varieties exist between H and L, blurring the boundaries of Ferguson's original dichotomy. Despite pressures from globalization and language contact, Arabic diglossia remains a defining characteristic of the Arab world, deeply intertwined with issues of identity, education, and cultural heritage.

In our work, the term *digital diglossia* is defined as a means of practice on social media platforms where different varieties of the same language strategically deployed, as it is the case with Arabic diglossia used in digital communication (see also Alkhamees et al. 2019). In this study, we aim to examine which variety of Arabic dominates Arabic discourse across X and Facebook. Arabic diglossia has also been examined from perspectives that extend beyond the traditional distinction between High and Low varieties. Bassiouney (2009), for example, provides a broad sociolinguistic account of Arabic in relation to language variation, identity, and social context. More recent research has examined the implications of diglossia for language development, literacy, and education. Khamis-Dakwar and Froud (2019) discuss the relationship between diglossia and language development, while Froud and Khamis-Dakwar (2021) critically review research on Arabic language acquisition in a diglossic context. Similarly, Khamis-Dakwar et al. (2022) examine diglossic knowledge and awareness in relation to language and literacy, and a recent systematic review highlights the continuing importance of diglossia in Arabic teaching and learning (Thomure et al. 2025). These perspectives provide an important foundation for examining how Arabic diglossia is manifested in newer digital communicative environments. Arabic digital diglossia forms part of this ongoing debate, particularly across X and Facebook, where distinct communicative norms may shape language practices. While X emphasizes brevity, immediacy, and rapid interaction, Facebook allows for longer and more elaborate discourse. With the rapid development of artificial intelligence (AI) and digital technologies, social media discourse has also evolved considerably (cf. Shormani 2025). Social media platforms are therefore not neutral channels but rather (re)shape discourse practices through their technological constraints and affordances (Androutsopoulos 2014; Herring 2019). Thus, the two platforms can be viewed as complementary contexts for examining Arabic digital diglossic discourse.

In Arabic context, social media has played a prominent visible role in shaping public discourse, most notably during events such as the Arab Spring, where platforms like X provided a space for mobilization and expression (Tudoroiu 2014). Nowadays, they provide a room for Arabs to express their views regarding what is going on in Palestine (see also Abusheikh 2023). However, cross-platform comparative work remains scarce, leaving open questions about how discourse practices vary between X and Facebook, and more importantly, how do Arab Facebook users and X users express their thoughts in Arabic, do they favor SA or CA?

## 2.2. Corpus linguistics

Corpus linguistics plays a substantial role in studying language through large, structured collections of texts known as corpora (Kennedy 2014). It provides a systematic, empirical approach to analyzing linguistic patterns by relying on real-life usage rather than introspection alone. Analyzing corpus data has been long established (Biber 1990). Through computational tools and quantitative methods, corpus linguistics enables researchers to investigate frequencies, collocations, concordances, and semantic or pragmatic trends across genres, registers, and dialects (Thompson 2014; McEnery and Brezina 2022; Wu 2023). Put differently, beyond describing language as it is used, corpus-based research supports applied fields such as lexicography, discourse analysis, sociolinguistics, and translation studies (Kennedy 2014; Giannossa 2015). Its strength lies in combining quantitative analysis with qualitative interpretation, allowing linguists to identify both broad tendencies and subtle variations in natural language use. Additionally, corpus linguistics offers a systematic and replicable framework for studying social media discourse (Laitinen and Rautionaho 2025; Sardinha 2022). Recently, corpus methods have been increasingly applied to digital communication, demonstrating how computational tools can reveal pragmatic markers, stance-taking, and identity strategies in online interaction. Liu (2020) demonstrates the effectiveness of corpus-based approaches in CDA, illustrating how quantitative and qualitative analyses of digital texts can uncover systematic patterns of power, ideology, and representation embedded in online discourse. The findings underscore the critical role of social media as both a mirror and a constructor of national images, revealing the interplay between language, ideology, and public perception in digital environments.

For Arabic, corpus-based studies remain relatively limited compared to English or other languages, though there are some studies utilizing Twitter corpora, online forums, and comment threads (Abusheikh 2023; Alqulaity et al. 2024; Alsuwaylimi 2024, see also Alnawas and Arıcı 2018, for a critical review), and Facebook (Banikhalef and Rababah 2018), but the scope of these studies is not (Arabic) diglossia. These works show that Arabic digital corpora present unique challenges, including orthographic variation, script mixing, dialectal diversity, and thematic discourse. This highlights the importance of using corpus-based approaches in studying Arabic discourse across digital platforms, where variation is not merely linguistic, but also socially and discoursally motivated.

## 2.3. Social media scraping

Python functions as both a data collection tool for crawling and scraping public posts, and as a screening mechanism to structure and clean the corpus prior to linguistic analysis. For instance, Catanese et al. (2011) employed Python web crawlers that extract Facebook data, choosing Python because of its portability, simplicity, and the wide range of libraries available for handling web requests, HTML parsing, and graph construction. The crawler mimicked user navigation of Facebook, systematically visiting public profiles, extracting friendship links, and storing these in anonymized graph structures. Python's ability to interface easily with databases and graph libraries also allows scholars to transform raw data into undirected graphs suitable for social network analysis. In this study, Python functions as both a data collection tool via automated crawling and scraping of public Facebook data, and as a screening tool of the data, thus structuring and cleaning the dataset before network analysis (cf. also Steinert-Threlkeld 2018; Mancosu and Vegetti 2020).

Regarding X, Python provides easy access to libraries such as Tweepy via X API to extract data from X. For instance, Graff et al. (2022) constructed a Python library "text_models" designed to streamline exploratory data analysis on X by leveraging token frequencies and aggregated origin–destination information from December 2015 onwards. It was developed as an open-source package, enabling them to extract daily counts of words and bigrams across four major languages, viz., Arabic, English, Spanish, and Russian, capturing user mobility trends across more than 200 countries and territories. Using the X's (Twitter) public API, the library retrieves tweets with or without geotags and preprocesses them into analyzable formats, facilitating tasks such as event mining, dialect comparisons, topic modeling, and word-cloud generation. Additionally, mobility reports produced by the library have been validated through correlation with Google's mobility data, demonstrating its reliability in capturing movement patterns. It illustrates multiple applications, from linguistic variation and semantic analyses to population mobility monitoring, offering a practical, high-level toolkit for mining social phenomena via X data.

### 2.4. Digital discourse analysis

Besides its conversational nature, discourse can also be understood as 'text' comprising linguistic and sociocultural elements. For decades, Hallidayan linguistics has been widely used as a method for analyzing texts. CDA views discourse as a "theoretical nexus," integrating multiple frameworks. Central to CDA is the analysis of linguistic features, especially syntax and semantics, which are crucial for understanding social meaning (Fairclough 2003). However, in DA, Yu and Chang (2024) argue, CDA involves considerable challenges due perhaps to its "mediated texts which are more interactive, intertextual, and heteroglossic do not share the same texture with traditional texts" (p. 460). Thus, the developments technology and internet have witnessed result in a digital discourse analysis (DDA) (cf. Shormani 2025). DDA is an approach to analyzing digital data in the digital age. Traditional CDA methods (Halliday 1994; Fairclough 1992, 2003) may not fit this age. In this age of DDA, traditional methods could be updated to fit the types and amounts of texts involved (Yu and Chang 2024). Along this line, Karlsen (2023) presents a new conceptualization of Foucault's archeological discourse analysis with digital methodology, specifically discussing discourse on women prior to the first wave women's movement. She proposes that "a digitized Foucauldian discourse analysis is possible, using a combination of digital methodology and close reading" (p. 195). Additionally, Bach (2020) proposes a method to analyze corpus texts, namely the "sensorial discourses", stressing the need for an approach to analyzing such texts.

The digital age, with the changes it brought about, necessitates new methods and techniques to analyze social media discourse (Qian 2025). For example, Liu (2024) proposes incorporating "text mining" into CDA to deal with "corpus-assisted discourse", and applies it to the "digital image" of the relations between Hong Kong and China. Liu uses a computer-assisted text-mining tool and quantitative text mining and qualitative discourse analysis to study the way Hong Kong-ers prefer to image China at a number of discourse levels.

### 2.5. Literature review

Studies addressing Arabic diglossic discourse in social media outlets remain to some extent scarce. However, there are a number of studies that tackle different aspects in corpus-based discourses (Alsuwaylimi 2024; Alqulaity et al. 2024; Abdalhadi et al. 2023). For example, Abdalhadi et al. (2023) investigates how Jordanians used Facebook status updates during the COVID-19 pandemic, employing a corpus-based pragmatic approach. These researchers compiled a corpus of Facebook posts written in Arabic during different phases of the pandemic and analyzed them for pragmatic functions such as expressing emotions, sharing information, giving advice, and promoting solidarity. Findings show that Facebook served as both an informational platform and a social support space. Users frequently employed speech acts such as directives, urging compliance with health measures, expressives, sharing fear, frustration, or hope, and commissives, viz., promising to follow rules. Pragmatic markers included emojis, repetition, humor, and religious expressions, which helped mitigate anxiety and reinforce community bonds. They conclude that Facebook discourse in Jordan during COVID-19 reflects both the diglossic nature of Arabic, mixing dialects with SA, and pragmatic strategies that emphasize solidarity, reassurance, and resilience. It concludes that social media played a crucial role in shaping public discourse and social interaction during times of crisis.

Another research has been conducted by Ilyas and Khushi (2012) who apply speech act theory to analyze Facebook status updates as acts of communication rather than mere statements. Using a dataset of status updates collected from Facebook users, they categorized posts according to speech act types: representatives/expressives, assertives, directives, commissives (cf. Searle 1969). The analysis found that expressive acts including sharing feelings, personal moods, or emotional states, were the most frequent, followed by directives. Commissives and declarations appeared less often, but still played roles in showing commitment such as promises, and marking social events including birthdays and achievements. They conclude that Facebook statuses are pragmatic tools for identity expression and social interaction, reflecting how users construct and negotiate relationships online. Additionally, Liu (2020) investigates how social media both constructs and reflects national images through a corpus-based CDA. Liu compiles a sizable corpus of social media posts from platforms where users actively discuss political, cultural, and social aspects of different nations. By systematically examining linguistic and discursive patterns, Liu identifies recurring themes, lexical choices, and evaluative language that contribute to the construction of national identity and influence public perception. The analysis identifies how users employ specific discursive strategies such as repetition, metaphor, nominalization, and evaluative adjectives, to promote, criticize or negotiate the image of a nation. Specifically, Liu's work concludes that digital area functions as a site where ideological positions are enacted, contested, and disseminated, allowing users to collectively shape and reshape narratives about national identity.

Concerning research at the Arabic level, there are some studies, though different from ours in scope, purpose, and methodology. For example, Alshutayri and Atwell (2019) present the development of a social media corpus of Arabic dialect texts, compiled from user-generated content across various platforms. The corpus captures the diversity of Arabic dialects and reflects the informal nature of online communication. They highlight challenges in working with Arabic social media text, including spelling variation, code-switching, non-SA orthography, and mixing SA with dialects. The authors also describe methodological approaches for corpus construction, including data collection, cleaning, annotation, and metadata management. Overall, this work provides a valuable resource for linguistic, sociolinguistic, and computational analyses of Arabic social media discourse. Additionally, Abusheikh (2023) examines digital activism related to Palestine on Twitter, comparing Arabic and English tweets using a corpus-based approach. The study analyzes linguistic, pragmatic, and discursive strategies employed in both languages, focusing on features such as hashtags, lexical choices, emotive expressions, and rhetorical structures. The findings reveal that Arabic tweets tend to be more emotionally expressive and culturally grounded, while English tweets often prioritize informational clarity and persuasive framing (cf. also Aljasir 2020; Omar and Ilyas 2018).

An important feature of Arabic diglossia is the considerable linguistic distance between SA and CA, which extends beyond differences in vocabulary and pronunciation to aspects of grammar, morphology, and syntax. This distance has important implications for literacy because Arabic users acquire and use written Standard Arabic alongside spoken varieties that may differ substantially from the language of writing and formal literacy. Saiegh-Haddad and Henkin-Roitfarb (2014) highlight the linguistic and cognitive implications of the distance between Arabic dialects and SA, while Saiegh-Haddad and Spolsky (2014) discuss the challenges and prospects of acquiring literacy in a diglossic context. Thus, Arabic diglossia is relevant not only to spoken language use but also to writing, as users may negotiate between the conventions of Standard Arabic and features associated with spoken varieties. This distinction is particularly important in digital discourse, where written communication can accommodate both formal written conventions and features of everyday spoken language.

Regarding employing Python for scraping SMPs, as discussed above, we followed Catanese et al.'s (2011) methodology. They address the problem of accessing large-scale social network data from Facebook, where privacy restrictions and technical barriers limit research opportunities, by developing privacy-compliant crawlers capable of collecting massive anonymized datasets of user connections. We also adopted methods from Graff et al. (2023) who have also used Python library as a tool for extracting data from SMPs such as X, as it facilitates exploratory analysis of X data by combining token-based text analysis with aggregated origin, destination information.

## 3. Methodology

### 3.1 Abiding by SMPs' terms of use

We abide by terms of use of both X and Facebook, and every possible action was taken to maintain adherence to terms of use and to safeguard user privacy, and ethical concerns. A large corpus of 16754 tweets and posts were collected using the Python-based libraries like *Tweepy*, adopting some methods from (Catanese et al. 2011; Graff et al. 2023). The posts selected are written in Arabic, publicly accessible, and from public pages on both SMPs to ensure ethical compliance, in that these pages are possible to extract data from, abiding by both SMPs' terms and conditions of use. Thus, we do not involve personal pages; we include only pages that are public by default and are designed for news, science, technology, businesses, brands, and public figures. For example, the X accounts include Aljazeera, AlArabiya, Russia Today, and China in Arabic, CNNArabic, Culture without limits, BBCArabic, Alekhbariya Sport. We have also included non institutional pages such as Gorgeous, Wada7 which are almost CA-oriented. Our aim to include these institutional and public pages was twofold: i) aiming to examine widely circulated public discourse on both platforms, and ii) to abide by both SMPs' user privacy, as was just noted. Both Facebook and X corpora consist of 5000 tweets/posts each, after irrelevant data were excluded. They were carefully balanced to provide comparable coverage of topics and discourse types. We also ensured that our combined corpus represents a mix of SA and CA.

### 3.2 Data collection, tools and sources

To collect our data, Python-based tools were used for both X and Facebook, focusing exclusively on publicly available content. For X, we employed *Tweepy* through which we accessed X API to extract posts from public accounts. We collected the data on 7 June, 2024. The timeframe was set from 01-01-2010 to 31-12-2024. Our aim was to collect data after launching both SMPs, which was the beginning of digital diglossia (both SMPs were almost launched at the same year, precisely 2006). In our data, only the text content of each post was collected, while engagement metrics such as likes, retweets, shares, or comments were excluded. This approach ensures a corpus of authentic text. Fig 1 presents part of the Python code script used to extract data from X via *Tweepy*.

```
File  Edit  Shell  Debug  Options  Window  Help
Python 3.10.5 (tags/v3.10.5:f377153, Jun  6 2022, 16:14:13) [MSC v.1929 64 bit (
AMD64)] on win32
Type "help", "copyright", "credits" or "license()" for more information.
>>> import tweepy
... import pandas as pd
... import time
... from datetime import datetime, timezone
...
... bearer_token = "[REDACTED]"
... api_key = "[REDACTED]"
... api_secret = "[REDACTED]"
... access_token = "[REDACTED]"
... access_token_secret = "[REDACTED]"
...
... client = tweepy.Client(
...     bearer_token=bearer_token,
...     consumer_key=api_key,
...     consumer_secret=api_secret,
...     access_token=access_token,
...     access_token_secret=access_token_secret,
...     wait_on_rate_limit=True
... )
...
... usernames = ["AJArabic", "alarabiy", "Yamenel1", "mog_china", "BBCArabic", "cnna
rabic", "KWBShj"]
... limit_per_user = 1000
... all_tweets = []
...
... start_time = datetime(2010, 01, 01, tzinfo=timezone.utc)
... end_time = datetime(2024, 12, 31, tzinfo=timezone.utc)
```

***Fig 1: Part of Python code script (X)***

All posts from both SMPs were saved as Comma-Separated Values (CSV) files for subsequent analysis, including manual classification into major discourse themes such as *politics, culture, technology, science,* and *business*. We examine 40 tweets/posts, 10 posts from each source to group them thematically, revealing differences between X and Facebook in terms of these 7 categories: *politics culture, technology, science, business, sports* and *fun* (see Table 2 & 3).

### 3.3. Data sampling and refinement

We collected 16754 X and Facebook posts, but only 10000 posts were included in our analysis after removing duplicate, non-Arabic, and irrelevant posts. We chose 10000 tweets to make our data balanced between both SMPs, and avoid H/L bias. Table 1 displays data distribution in terms of study data, duplicates, and cades, ambiguous, empty, unused tweets/posts, Freq and percentages

**Table 1: Total and net data**

| Platform | Net data | | Duplicates | | Hard/amb. cases | | Empty | | Inappropriate | | Unused | | Total | |
|---|---|---|---|---|---|---|---|---|---|---|---|---|---|---|
| | N | % | N | % | N | % | N | % | N | % | N | % | N | % |
| **X** | 5000 | 59.84 | 1374 | 16.44 | 463 | 5.54 | 19 | 0.23 | 902 | 10.80 | 597 | 7.15 | 8355 | 100 |
| **Facebook** | 5000 | 59.53 | 1417 | 16.87 | 1720 | 20.48 | 52 | 0.62 | 7 | 0.11 | 201 | 2.39 | 8399 | 100 |
| **Total** | 10000 | 59.69 | 2791 | 16.66 | 2183 | 13.03 | 71 | 0.42 | 911 | 5.44 | 798 | 4.76 | 16754 | 100 |

Table 1 outlines the data filtration pipeline used to derive the final analytical sample of 10,000 posts from an initial collection of 16754 raw entries across X (*n* = 8355) and Facebook (*n* = 8399). Exactly 5000 clean entries (59.69% overall yield) were retained per platform after removing invalid posts. The primary sources of exclusion were duplicate posts (16.66%) and hard/ambiguous linguistic cases (13.03%). Notably, X exhibited a higher rate of inappropriate content (10.80% vs. 0.11% on Facebook), whereas Facebook yielded a larger proportion of ambiguous language cases (20.48% vs. 5.54% on X). Remaining exclusions comprised empty posts (0.42%) and unused entries (4.76%), resulting in a balanced, high-quality corpus for downstream diglossic analysis.

In our sampling, a subset of 40 tweets/posts was selected from the full dataset of 10000 items. The selection followed a systematic sampling procedure, whereby posts were extracted at regular intervals across the dataset, i.e. posts 997-1000, 1997-2000, and so on. This approach ensured coverage of the entire dataset and minimized selection bias, which was meant to illustrate the qualitative dimension of the analysis. It is important to note that these examples are not intended to be statistically representative, but rather to serve as illustrative instances supporting the broader quantitative findings. This is consistent with established practices in discourse analysis, where selected excerpts are used to exemplify patterns identified in larger datasets (see e.g., Talja 1999).

After collecting the raw tweets/posts from X and Facebook, the dataset underwent a rigorous data refinement process to ensure quality, relevance, and consistency for the analysis we performed. Our all dataset contained 16754 posts/tweets in total, 2791 of which were duplicates which were identified and removed to avoid redundancy. Posts containing non-Arabic text, advertisements, or spam were manually filtered out to maintain linguistic focus on Arabic digital diglossic discourse. The dataset was also manually refined, hence, removing symbols including &, *, @, ^, Persian numbers including “. ١, ٢, ٣…”, emojis, underscores “_”.We excluded hashtags, though being a salient ingredient of digital discourse, because including them would reveal users’ personal information and accounts and this violates our criteria of not including personal information to abide by both SMPs’ user privacy. However, we did not exclude English words/phrases aiming to show code-switching between Arabic and other languages, specifically English (see Table 6). This refined dataset was then used for dialect classification, allowing for posts to be grouped into varieties such as SA CA, and categories including politics, culture, science. We aim to ensure that the corpus was clean, clearly classified, reliable, and suitable for subsequent qualitative and quantitative analyses of Arabic diglossic discourse across both SMPs.

### 3.4. Data annotation and classification

The dataset was annotated manually through 2 phases: phase 1 was conducted by 2 PhD students of linguistics; phase 2 was conducted by the authors in which we reevaluated what the coders produced to ensure consistency. The training process was conducted for 2 weeks involving the authors who trained the 2 coders, independently annotating a random subset of 2000 tweets/posts per platform for SA and CA. SA and CA are identified through a number of criteria including orthography, lexicon, morphology, syntax, and discourse markers (Shormani and Al Hussen 2024). SA is commonly characterized by standardization of these features, while CA is identified through dialectal lexicon, non-standard linguistic features, and dialect-specific discourse markers. For example, جاء المشاركون يوم أمس وقد بعث رئيس المجلس رسالته صباحاً, versus المشاركين جم وكان ريس المجلس بعت رسالته امبارح بدري, versus المشاركين جم وكان ريس المجلس بعت رسالته امبارح بدري '*the participants came, and the chairman of the council had sent his message yesterday early morning'*, we considered the first a SA sentence, and the second a CA one, base distinctive features of CA, specifically Egyptian Arabic. Regarding orthography: رئيس vs. ريس, 'president', SA uses ئ while CA does not use it. another example is بعث vs بعت). As for lexicon: يوم أمس vs امبارح 'yesterday', يوم أمس is a SA standardized phrase while امبارح is a CA, (another example is صباحاً vs بدري). Morphology and syntax also appear as in المشاركون vs. المشاركين ending in ون vs ين, the former indicates Nominative case because it a subject while the latter Accusative Case is used to mark objects (Shormani and Qarabesh 2018). However, in the CA sentence, we find المشاركين though it is a subject. CA varieties lose this distinctive feature of SA.

Mixed registers/posts were categorized based on dominant features, following established practices in Arabic sociolinguistics. In Arabic sociolinguistics, specifically in studies of diglossia, texts are often not purely SA or purely CA (see e.g., Bassiouney 2009). Therefore, a post was labeled SA or CA depending on which linguistic features are dominant, not whether it is "pure." For example, نناقش اليوم تأثير التكنولوجيا على التعليم بس لازم نكون واقعيين '*we discuss today the impact of technology on education, but we need to be realistic*'. This post was labelled as SA though it contains the dialectal term "بس" 'but' for the SA "لكن", and "لازم" 'necessary'. Here, for instance, we considered the number of words, belonging to SA, syntax "نكون واقعيين", VSO rather than SVO. However, posts like التكنولوجيا مهمه بس ها اليومين الناس ما تستخدمها صح '*technology is important, but these days people do not use it properly*' were labelled as CA.

After the data annotation, we classified the net data, 10000 tweets/posts into SA or CA. To ensure the reliability of the SA/CA classification, inter-coder agreement was assessed on a randomly selected subset of 2000 tweets/posts. 2 coders independently annotated each item as either SA or CA. Cohen's kappa (κ) was calculated using Python's sk-learn library, yielding $\kappa = 0.79$, which indicates substantial agreement beyond chance (cf. Landis and Koch 1977). The sampling procedure was made reproducible through the use of a fixed random seed ($n = 42$), ensuring that the same subset can be retrieved in future analyses.

### 3.5. Methods of Analysis

As noted so far, the digital age comes with new forces and changes to the traditional approaches and methods, ending up with new (modified) approaches, suiting these forces and changes. Thus, adopting methods from works by (Fairclough 2003; Yu and Chang 2024; Bach 2020; Qian 2025; Liu 2024) we propose that in DDA research an approach could look at three levels of analysis: i) micro-level, a linguistic level, ii) macro level, a level which tackles the sociocultural aspects, and iii) digital level, a level that may focus on the forces, effects and consequences of SMPs on discourse. Thus, in our context the choice of words in Arabic, be it SA or CA, signals how users represent, construct, and interpret social realities on X and Facebook. We also propose that DDA should see digital texts as a result of both "social practice", and digital practice analyzing and interpreting the sociocultural aspects, and the digital forces shaping language use, here diglossia. A DDA approach to Arabic diglossia includes how diglossic practices reflect, maintain, or challenge social norms, ideologies, and power structures within Arabic-digital spaces, here X and Facebook. This amalgamated approach provides a comprehensive understanding of ADD on both SMPs, highlighting both quantitative trends and qualitative nuances.

Additionally, to examine whether the associations of SA and CA across discourse categories is statistically significant, bivariate and logistic regression analyses were conducted using Python's *scipy* library. Bivariate analyses were used to examine the initial associations between platform, discourse category, and diglossic choice, with effect sizes reported where appropriate. A binary logistic regression model with Platform × Discourse Category interaction terms was then fitted to assess whether the association between discourse domain and diglossic choice varied across platforms.

## 4. Results

This section presents the study results from the two sources, viz., X and Facebook. We have divided the data into 7 categories, namely *politics, culture, technology, science, business*, *fun,* and *sports*. We have also classified each category into SA and CA. Table 1 presents data from X and Table 2 displays Facebook data.

**Table 2: Distribution of diglossic & discourse category on X**

| **Diglossic Cat.** | **SA** | | **CA** | | **Total *N*** | **%** |
|---|---|---|---|---|---|---|
| **Discourse Cat.** | ***N*** | **%** | ***N*** | **%** | | |
| **Politics** | 1201 | 75% | 391 | 25% | 1592 | 31 |
| **Culture** | 689 | 63% | 406 | 37% | 1095 | 22 |
| **Technology** | 437 | 67% | 215 | 33% | 652 | 13 |
| **Science** | 335 | 57% | 259 | 43% | 594 | 12 |
| **Business** | 245 | 47% | 280 | 53% | 525 | 11 |
| **Fun** | 61 | 14% | 372 | 86% | 433 | 9 |
| **Sports** | 65 | 60% | 44 | 40% | 109 | 2 |
| **Total** | 3033 | 60.7% | 1967 | 39.3% | 5000 | 100 |

**Table 3: Distribution of diglossic & discourse category on Facebook**

| Diglossic Cat. | SA | | CA | | Total *N* | % |
|---|---|---|---|---|---|---|
| Discourse Cat. | *N* | % | *N* | % | | |
| **Politics** | 711 | 70% | 303 | 30% | 1014 | 20 |
| **Culture** | 451 | 32% | 923 | 68% | 1374 | 28 |
| **Technology** | 310 | 58% | 142 | 42% | 452 | 9 |
| **Science** | 510 | 71% | 211 | 29% | 721 | 14 |
| **Business** | 231 | 36% | 412 | 64% | 643 | 13 |
| **Fun** | 14 | 2% | 708 | 98% | 722 | 14 |
| **Sports** | 3 | 13% | 71 | 87% | 74 | 2 |
| **Total** | 2230 | 55.4% | 2770 | 44.6% | 5000 | 100% |

Tables 2 and 3 present the distribution of SA and CA across discourse categories on X and Facebook. On X, SA clearly dominates, accounting for 3033 tweets, 60.7%, compared to 1967 tweets, 39.3% in CA, out of a total of 5000 tweets. On Facebook, however, the pattern reverses: CA is more frequent, with 2770 posts, 55.4%, while SA accounts for 2230 posts, 44.6%. Across the combined dataset, $n$ = 10000, SA constitutes 5263 posts, 52.6%, while CA represents 4737 posts, 47.4%, indicating a slight overall preference for SA but with strong platform-based variation. At the level of discourse categories involved in our study, *politics* is the most prominent category on X, with 1592 tweets, 31. % of the platform total. Within this category, SA is overwhelmingly dominant, 75%, $n$ = 1201, compared to CA 25%, $n$ = 391. On Facebook, *politics* accounts for 1014 posts, again showing strong SA dominance 70%, $n$ = 711 over CA 30%, $n$ = 303. These figures confirm that political discourse is consistently associated with SA across both platforms.

However, *culture* emerges as the dominant category on Facebook, with 1374 posts, where CA clearly prevails 68%, $n$ = 923 over SA 32%, $n$ = 451. On X, *culture* represents 1095 tweets, but shows a different pattern, with SA at 63%, $n$ = 689 and CA at 37%, $n$ = 406. This indicates that while cultural discourse on Facebook is strongly colloquial, it remains more formalized on X. *Technology* and *science* display relatively similar distributions across platforms, though with a consistent preference for SA. On X, *technology* accounts for 652 tweets, with SA at 67% $n$ = 437 and CA at 33%, $n$ = 215, while *science* comprises 594 tweets 11.9%, with SA at 57%, $n$ = 335 and CA at 43% $n$ = 259. On Facebook, *technology* represents 452 posts, with SA at 58%, $n$ = 310 and CA at 42%, $n$ = 142, whereas *science* accounts for 721 posts, showing stronger SA dominance 71%, $n$ = 510 compared to CA 29%, $n$ = 211. These figures confirm that knowledge-based discourse tends to favor SA, particularly on Facebook in the case of *science*.

The categories *fun* and *sports* show the strongest association with CA, specifically on Facebook. On X, *fun* constitutes 433 tweets, with a clear CA dominance 86%, $n$ = 372 over SA 14%, $n$ = 61, while *sports* is relatively minor 109 tweets, with SA at 60%, $n$ = 65 and CA at 40%, $n$ = 44. On Facebook, entertainment represents 722 posts, and is almost exclusively CA 98%, $n$ = 708, with minimal SA presence 2%, $n$ = 14. Similarly, *sports*, though small in scale 74 posts; 2%, is dominated by CA 87%, $n$ = 71 compared to SA 13%, $n$ = 3. These patterns highlight the strong link between informal, entertainment-oriented discourse and CA, particularly on Facebook. Finally, *business* discourse presents a mixed pattern. On X, it accounts for 525 tweets, with a slight CA majority 53%, $n$ = 280 over SA 47%, $n$ = 245. On Facebook, *business* comprises 643 posts, where CA is more clearly dominant 64%, $n$ = 412 compared to SA 36%, $n$ = 231. This suggests that commercial discourse, particularly on Facebook, tends to favor a more informal, accessible linguistic style.

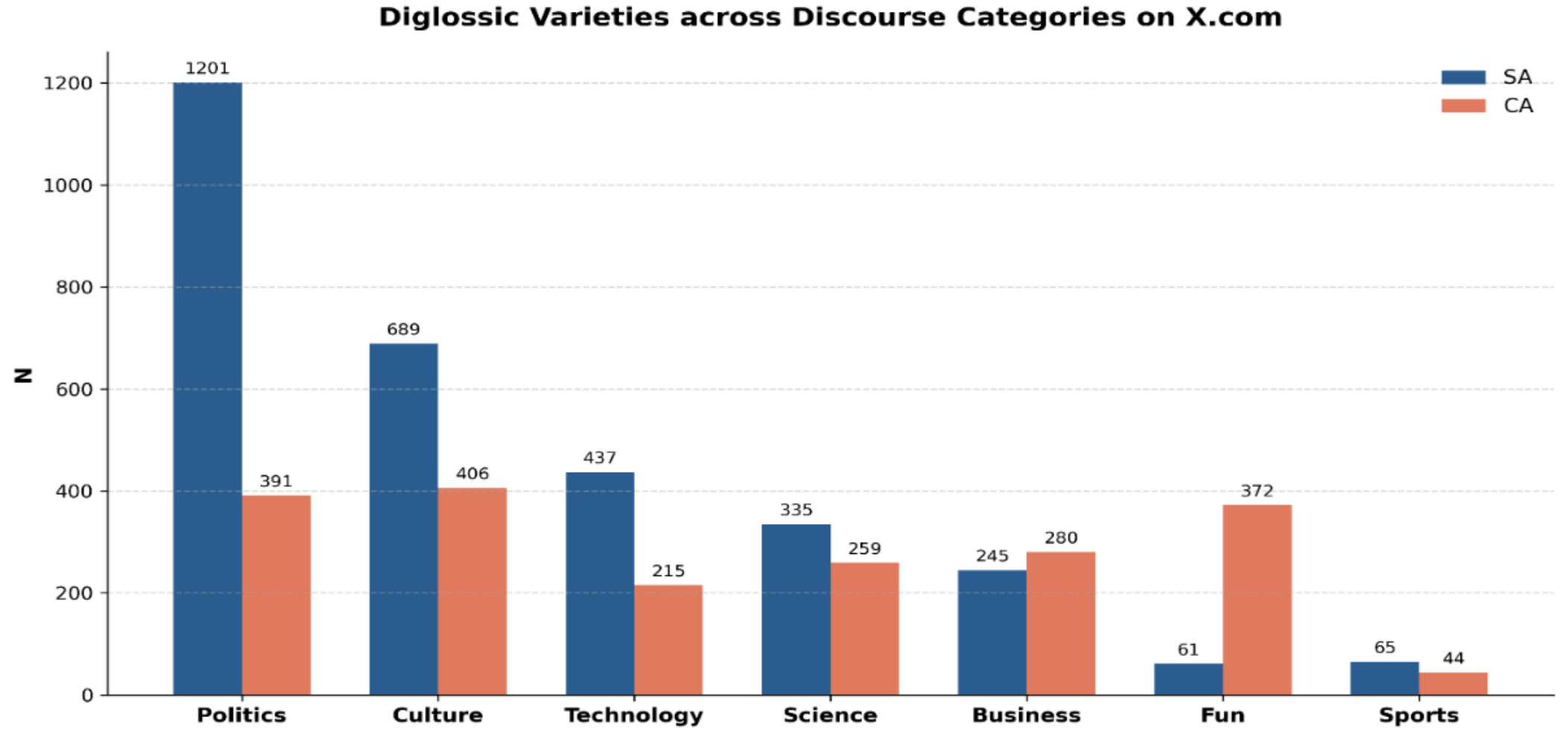

***Fig 2: Diglossia vs discourse on X***

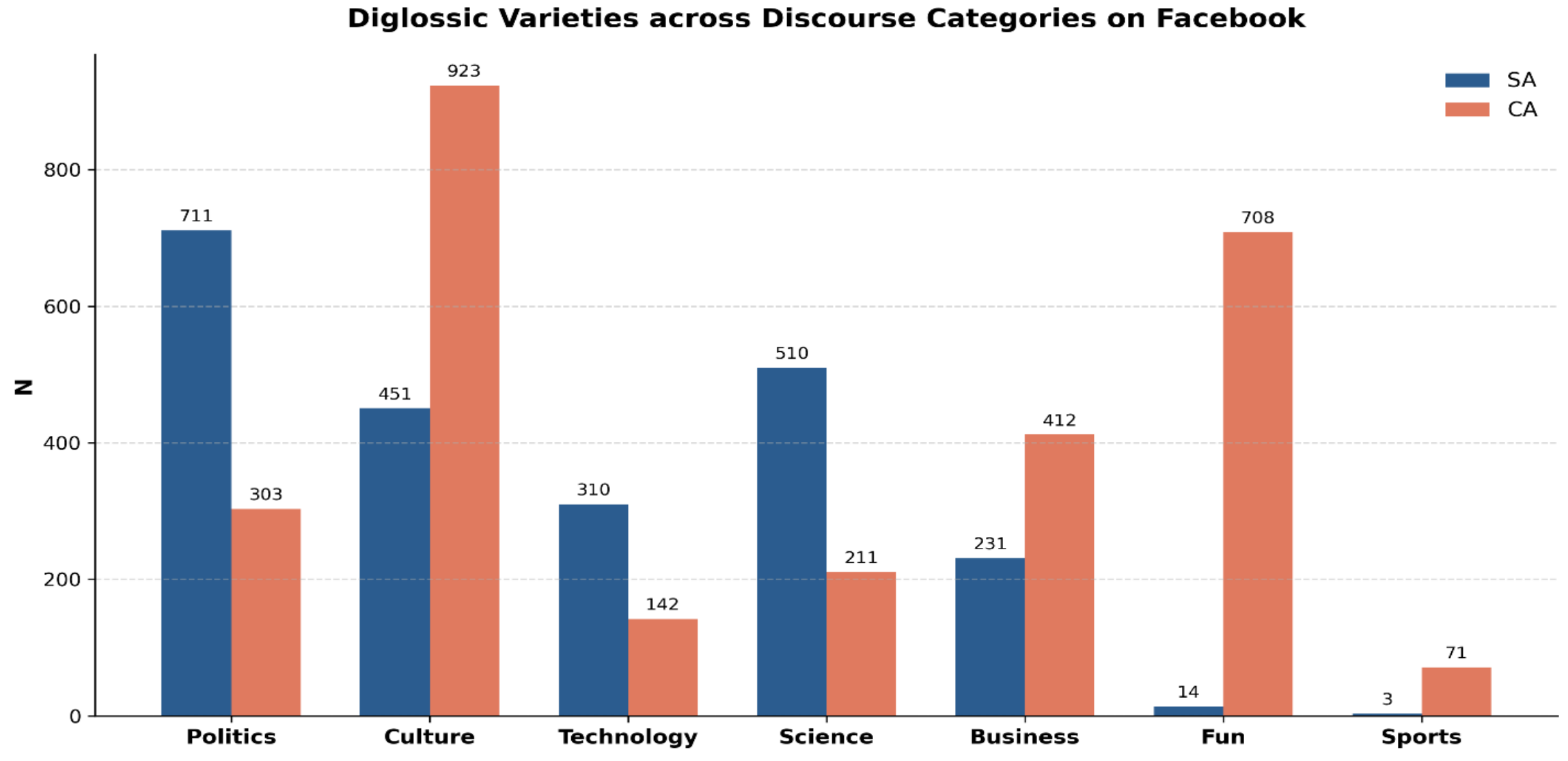

*Fig 2: Diglossia vs discourse on Facebook*

**Table 4: Bivariate relationships across discourse category, SMPs, and diglossic choice**

| Analysis Level | Relationship tested | χ2 | df | p | CV |
|---|---|---|---|---|---|
| **X** | Discourse Domain x Diglossic Choice | 600.35 | 6 | .001 | .347 |
| **Facebook** | Discourse Domain x Diglossic Choice | 1249.52 | 6 | .001 | .500 |
| **Pooled** | Discourse Domain x Diglossic Choice | 1883.03 | 6 | .001 | .434 |
| **Cross-Platform** | Platform x Diglossic Choice | 262.16 | 1 | .001 | .162 |
| **Cross-Platform** | Platform x Discourse Domain | 133.50 | 6 | .001 | .115 |

Table 4 To evaluate the bivariate relationships between social media platforms, discourse domains, and language variety choices, Chi-square tests of independence were conducted alongside CV to determine effect sizes.

To examine the distribution of SA and CA within each platform, separate 7 × 2 Chi-square tests were performed. On X, language choice varied significantly across discourse domains, $\chi^2(6, n = 5000) = 600.35$, $p < .001$, demonstrating a moderate-to-strong association ($CV = .347$). On Facebook, the relationship between discourse domain and diglossic choice was even more pronounced, $\chi^2(6, n = 5000) = 1249.52$, $p < .001$, yielding a strong effect size ($CV = .500$). When pooling the dataset across both platforms ($n = 10000$), the overall association between discourse domain and diglossic choice remained highly significant, $\chi^2(6, n = 10000) = 1883.03$, $p < .001$, $CV = .434$.

Cross-platform main effects were also evaluated. A 2 × 2 test confirmed a statistically significant overall association between platform type and diglossic choice, $\chi^2(1, n = 10000) = 262.16$, $p < .001$, though the effect size was relatively weak ($CV = .162$), reflecting higher baseline SA usage on X compared to Facebook. Finally, a 7 × 2 test indicated a significant relationship between platform type and topic distribution, $\chi^2(6, n = 10000) = 133.50$, $p < .001$, $CV = .115$, confirming that the volume of posts across discourse domains differed between the two platforms.

**Table 5: Binary logistic regression for SA choice**

| Predictor / Term | Coef. (B) | Std. Err. | z-value | p | Odd Ratio | 2.5% CI | 97.5% CI |
|---|---|---|---|---|---|---|---|
| Intercept (FB, Politics) | 0.8529 | 0.0686 | 12.4324 | 0.001 | 2.3465 | 2.0513 | 2.6843 |
| X vs. FB | 0.2693 | 0.0900 | 2.9924 | 0.0028 | 1.3090 | 1.0974 | 1.5615 |
| Business | −1.4315 | 0.1071 | −13.3708 | 0.001 | 0.2389 | 0.1937 | 0.2947 |
| Culture | −1.5691 | 0.0895 | −17.5349 | 0.001 | 0.2082 | 0.1747 | 0.2482 |
| Fun | −4.7763 | 0.2785 | −17.1518 | 0.001 | 0.0084 | 0.0049 | 0.0145 |
| Science | 0.0296 | 0.1068 | 0.2773 | 0.7816 | 1.0301 | 0.8355 | 1.2699 |
| Sports | −4.0170 | 0.5934 | −6.7695 | 0.001 | 0.0180 | 0.0056 | 0.0576 |
| Technology | −0.0722 | 0.1224 | −0.5900 | 0.5552 | 0.9304 | 0.7320 | 1.1825 |
| X × Business | 0.1758 | 0.1500 | 1.1719 | 0.2412 | 1.1922 | 0.8885 | 1.5997 |
| X × Culture | 0.9758 | 0.1237 | 7.8856 | 0.001 | 2.6533 | 2.0819 | 3.3815 |
| X × Fun | 1.8461 | 0.3163 | 5.8373 | 0.001 | 6.3351 | 3.4084 | 11.7748 |
| X × Science | −0.8945 | 0.1471 | −6.0803 | 0.001 | 0.4088 | 0.3064 | 0.5454 |
| X × Sports | 3.2850 | 0.6274 | 5.2359 | 0.001 | 26.7091 | 7.8094 | 91.3481 |
| X × Technology | −0.3407 | 0.1591 | −2.1419 | 0.0322 | 0.7113 | 0.5207 | 0.9715 |

Table 5 presents the binary logistic regression predicting the choice of SA, coded as 1, vs CA, coded as 0, with Facebook and Politics serving as the reference categories. The intercept was significant (*B* = 0.8529, p < .001, *OR* = 2.3465). The main effect of platform was also significant: X was associated with higher odds of SA use than Facebook within the Politics reference category (*B* = 0.2693, p = .0028, *OR* = 1.3090, 95% CI [1.0974, 1.5615]). Relative to Politics on Facebook, the main effects of Business (B = −1.4315, p < .001, *OR* = 0.2389), Culture (*B* = −1.5691, p < .001, *OR* = 0.2082), Fun (B = −4.7763, p < .001, *OR* = 0.0084), and Sports (*B* = −4.0170, *p* < .001, *OR* = 0.0180) were significant and negative. Science (*B* = 0.0296, *p* = .7816, *OR* = 1.0301) and Technology (*B* = −0.0722, p = .5552, *OR* = 0.9304) were not significant.

The interaction terms indicate that the relationship between discourse category and SA choice varied by platform. Significant positive interactions were observed for Culture (*B* = 0.9758, p < .001, *OR* = 2.6533), Fun (*B* = 1.8461, *p* < .001, *OR* = 6.3351), and Sports (*B* = 3.2850, *p* < .001, OR = 26.7091). Significant negative interactions were found for Science (*B* = −0.8945, *p* < .001, OR = 0.4088) and Technology (*B* = −0.3407, p = .0322, *OR* = 0.7113). The Business interaction was not significant (*B* = 0.1758, *p* = .2412, *OR* = 1.1922). Thus, these interaction effects indicate that the association between discourse category and diglossic choice differs between X and Facebook for the categories with significant interaction terms.

## 5. Discussion

In this section, we present the qualitative analysis of the diglossic use of Arabic varieties, presenting examples from our corpus that represent the 7 categories in Tables 2 & 3. As Table 6 depicts, we present 10 examples for each diglossic variety on both platforms.

**Table 6: Digital diglossia: Arabic between X and Facebook (examples)**

| no | X | Facebook |
|---|---|---|

| | SA | CA | SA | CA |
|---|---|---|---|---|
| 1 | ليس هناك زواج كامل ولا صداقة كاملة ولا حياة وردية حرر عقلك من التوقعات الغير واقعية الجميع يُعاني | تنام عشان تنسى همومك وتتفاجئ ان الشيطان مجهز لك كابوس ملخص عن نفس الموضوع مع شوية دراما مع بهارات هنديه 🤣🤣 | بايدن: سياسة نتنياهو خاطئة وأدعو إسرائيل لوقف إطلاق النار | يلعب ديفون سوا دور القيادة الشجاع والاستطوائي، تحت إشراف لوك سبارك، الذي يجلب مشاهد الحركة الخلابة |
| 2 | على اقتراب أفول عام 2024تتجلى ملامح عام استثنائي ترك بصمات تاريخية غيّرت مسار الشرق_الأوسط .كان هذا العام شاهدًا على انتصارات حاسمة قادها جيش الدفاع لإسرائيلي الذي عزز ردعه وأثبت للجميع أن الافتراء عليه ثمنه باهظ جدًا | صباح الرضى وكل الأماني الطيبة صباح الفل والياسمين يارب سهّل لنا دروب الخير في كل إتجاه. | استعادة دور الحركة الجماهيرية ضرورة حتمية لإنهاء الحرب وفضح مصالح القوى المحلية والدولية الساعية إلى استمرارها بهدف تفكيك الدولة وإجهاض الثورة💪💪💪 . | جاااااالك تجارة اسكندرية او مستني المرحلة الثانية وعاوز تدخلها 🤣🤣 🤣 |
| 3 | سيُكتب على أبواب "غزّة" حكاية شعب انتصر، بعد أن تخلّى عنه معظم البش 💪 .. 🙏🙏🙏 | "ناس كتير هتقلع الفترة الجاية" شكلها ولعت، والجو حار؟، والأيام الجاية نار على زيت حار وفي نار الحر، ونار السياسية، الأيام الجاية، ناس كتير في العالم كله هتتزنق والقاعده بتقول :"كل ما تتزنق،.اقلع" وناس كتير هتقلع🤣🤣 | التداولية هي دراسة اللغة في سياق استعمالها الفعلي، وهي علم يبحث في المعنى التواصلي للغة من خلال دراسة العلاقة بين البنية اللغوية ومستعمليها والسياق الذي تُستخدم فيه. إنها تتجاوز المستوى التركيبي والدلالي المجرد للغة لتدرس كيفية إنتاج المعنى في المواقف التواصلية الحقيقية❤️ . | صرااااحة تعبتهم حيل خلال هذا الشهر اني بطبيعتي بالحياة اركز على التفاصيل و احسب حساب لكلشي و بنفس الوقت جدا غثيث باختيار شركة وساطة تناسب المتداولين و تناسب كل متابعيني بالمنطقة العربية و بالعراق خصوصاً مع كثرة شركات الوساطة بالفترة الاخيرة و المتداول صار محتار يا شركة وساطة ممكن يثق بيها او يتداول وياها . 🤣🤣🤣 |
| 4 | إما أن دولة المؤسسات ترغب بإكمال اجندا اليسار المتطرف، وبالتالي ستدفع بكامالا، أي باوباما، ليكمل ما بدأه في 2009 | إذا تبي ترفع جودة حياتك بشكل كبير جدا عندي لك نصايح بسيطة .العب رياضة يوم ايه ويوم لا .اكل وجبات تحت حساب السعرات المثالي ب 200 سعر حراري . تسامح مع اخطاء الجميع دايما ولا تاخذ اي شي بشكل شخصي إطلاقا . اجعل المسجد اجمل مكان في يومك .كن لطيفا مع الجميع بالذات والديك. | أجواء شديدة البرودة وصقيع.. تحذير عاجل بسبب الطقس غدا الخميس 26-12-❤️2024 | واحده قالت زوجي إتاخر على وم عندي رصيد مشيت لي حماتي قلتا ليها اديني تلفونك اتصل علي اخوك اشوفو مالو اها كتبتا رقمو لقيتا مسجلاهو بي (راجل القعونجه) ياخ بقيت لا بسمع ولا بشوف وهي شكلها اتذكرت الإسم جات جاريه لقتني خلصتا المكالمه وسالتني اها اتصلتي؟ قلت ليها غاق غاااااق غاااااق غاااااق 🤣 🤣 |
| 5 | SamSungتضرب بقوة Appleفي خطر أطلقت سامسونج هاتفها الجديد Galaxy AI S24والذي يأتي بدعم كبير للذكاء الاصطناعي💪 ❤️!! | الدجال سيخرج للعلن بعد أن تفشل مخططاته ! حين يظهر المهدي ، ويبدأ في استعادة الحكم من السيطرة الشيطانية ..لن يغوي الدجال أحدا ولن يخدعه ، سيتبعه | عُقد الإجتماع السابع لمجلس إدارة مصرف الواحة بمقر الإدارة العامة لمصرف الواحة باب تاجوراء -سوق الجمعة - طرابلس -ليبيا، بحضور السيد رئيس مجلس الإدارة والسادة الأعضاء، حيث تم مناقشة جدول | فراس قطوسي يتأهل إلى نصف نهائي التايكواندو عن فئة تحت 80 كلغ ضمن دورة الألعاب الأولمبية باريس 2024😍 😍 |

|  |  | المخدوعون به قبل أن يظهر! | الأعمال المطروح في هذه الجلسة وما يستجد من أعمال |  |
|---|---|---|---|---|
| 6 | عاجل فوز تاريخي وتأهل منتخب مصر إلي ربع نهائي اوليمبياد باريس 2024 بعد الفوز على إسبانيا وصيف أوروبا 2-1 Paris2024 | على كثر مازليت للحين فيني دين اسوي ذنوب كثار واصوم واصلى انا انسان واحد عاشوا بداخلي اثنين الاول مطوع .. لـكن الثاني مولي😭 😭 | قناة الجزيرة مباشر تنقل من البث المباشر للشبكة الإعلامية اليمنية الأمريكية المؤتمر الصحفي للجالية العربية الذي عقد في اليمن الصغرى في البرنكس - نيويورك لمؤيد المرشح الرئاسي للانتخابات الأمريكية 2024 ترمب | لكل المهتمين ستقام دورة الذكاء الاصطناعيAI اول محاضره مجانيه مع شهادة دولية مجانية من يريد الإنضمام👍 👍 |
| 7 | "أحب شعور أن أكون رفيقك الوحيد بالفرح، أن أكون ذلك الطريق الموصّل لراحتك، وأن أبقى ذلك السبب الذي يجعلك تبتسم كلّما مررتُ على بالك "."أحب شعور أن أكون رفيقك الوحيد بالفرح، أن أكون ذلك الطريق الموصّل لراحتك، وأن أبقى ذلك السبب الذي يجعلك تبتسم كلّما مررتُ على بالك." | ماااااشاءالله يمكن هذي اجمل قصة زواج بسنة 2024 مبرووووووك وفاء الشمري الاغنية تليق لك اجمل اغاني الحب جاب لها الخاتم من لندن و عقد القران بدبي و زواجها بيصير في لندن او باريس مدينة العشاق ❤️😍 ❤️ | تويوتا تاكوما 2024 الفخامة العربية لتجارة السيارات والمحركات | 2024 وضع الماء في الزعفرانيه يوم اكو و 10 ماكو 💔 💔 |
| 8 | الحمد لله على نعمة الصباح، الذي يأتي بجمال الأمل والتفاؤل، ويمنحنا فرصة جديدة لبناء ❤️❤️❤️أحلامنا وتحقيق أهدافنا🙏🙏 .. 🙏 | من فترة طورت نموذج لخطة التسويق الرقمي لعام 2024، تتكون من اربع مراحل وهي:😍 | الحسيني المدان بالتعامل مع الموساد تنبأ بمقتل نصر الله وحرم قتال الشيعة لإسرائيل. برز الأمين العام للمجلس الإسلامي العربي، محمد علي الحسيني | 2024كانت أصعب سنه مرت عليا في حياتي. وخسرت فيها كل الفلوس اللي في محفظتي الاستثمارية ، وخرجت من السنه دي خسران كل حاجه ، وطلعت منها يعلم ربنا بـ 10 آلاف جنيه آخر ما أملك من فلوس |
| 9 | .أتوقع أن تكون هذه آخر انتخابات أمريكية وأمريكيا على شكلها الحالي، ستتحول الولايات المتحدة إلى دويلات، سيسبقها شغب في الشوارع ودماء. | والله أنك في حياتي مثل همّال المطر تحيّي شعور المشاعر وتسقيهاأمل❤️ .. | مقابلة الأمير الحسين بن عبدالله الثاني، ولي_العهد، مع قناة العربية، بمناسبة "اليوبيل_الفضي "لتسلم جلالة الملك عبدالله الثاني سلطاته الدستورية | قبل_النوم اشتي أقول لكم خبر هام.. بعد بكره بإذن الله سيتم افتتاح محل جديد لإحدى الأخوات. بسبوسة_جود متخصص بإعداد وتقديم أحلى واشهى اطباق البسبوسة بأنواع مختلفة مع العصائر الساخنه. |
| 10 | لا تسمعوا ما أقول بل انظروا الى ...وكما الأرقام؟!! 💪 | تمام بسقيها واعتني فيها.. خلاااااص تصبحو ع خير | اذا كان الطب هو طب الاسنان، فالطب البيطري يعتني بالحيوان لا تضحكوا 🤣 🤣 🤣 | الرياض والدمام امطرت واحنا باقي، ترا وين الخاشعين يصلون بكره جمعه .. ج.م❤️❤️❤️ |

***Note: For Table 6 English translation, see Appendix***

## 5.1. Thematic analysis

Beyond the quantitative and digital discourse analyses, the dataset was examined thematically to identify recurrent discourse domains across X and Facebook and to explore how diglossic variation in SA vs. CA is distributed within these domains. The analysis reveals a consistent set of themes, including *politics, culture, science, business/advertising, technology, fun* and *sports*, with clear platform-based differences in both thematic prominence and linguistic realization. For example, political discourse emerges as a dominant theme on X, where posts are frequently articulated in SA and adopt a formal, authoritative tone. Examples such as " بايدن: سياسة نتنياهو خاطئة وأدعو إسرائيل لوقف إطلاق النار" *'Biden: Netanyahu's policy is wrong, and I call on Israel to cease fire'* and " إما أن دولة المؤسسات ترغب بإكمال اجندا اليسار المتطرف..." *'Either the state of institutions wants to continue the agenda of the radical left'* reflect institutional, ideological, and evaluative discourse, though more on X than on Facebook, in which the speaker reports Biden's political view towards the Middle East Cause, viz., the Palestine Cause (cf. Ekström et al. 2021). Similarly, Facebook also hosts political content, as in "استعادة دور الحركة الجماهيرية ضرورة حتمية...' *restoring the role of the mass movement is an inevitable necessity…*', yet such posts coexist with a wider range of informal and socially embedded themes. Across both platforms, politics is strongly associated with SA, reinforcing its role as the H variety used for authority, legitimacy, and pan-Arab communication.

However, cultural discourse is more prominent on Facebook and is predominantly expressed in CA. Posts such as "مااااشاءالله يمكن هذي اجمل قصة زواج..." 'Mashallah, perhaps this is the most beautiful marriage story…' and "2024 كانت أصعب سنه مرت عليا في حياتي..." '2024 was the hardest year I have ever experienced in my life…' illustrate how users employ CA to narrate personal experiences, emotions, and social relationships. These themes highlight Facebook's role as a space for identity construction, intimacy, and community bonding, where dialect functions as an index of authenticity and relatability. Another subtheme emerging from culture is what could be dubbed as morality and education discourse. This theme is embodied in Arabs' behavior, it also appears in both varieties, often blending SA and CA depending on the communicative intent. For example, tweets like "الحمد لله على نعمة الصباح..." *'Praise be to Allah for the blessing of the morning…'* employ SA to convey moral reflection and general wisdom, while advisory posts in CA " إذا تبي ترفع جودة حياتك..." 'If you want to improve your quality of life' reflect a more personalized and accessible tone. This thematic domain illustrates the fluidity of diglossic choices, where speakers shift between varieties to balance authority and intimacy.

Additionally, the entertainment domain, i.e. *fun* and *sports* further reinforces this pattern. Humor, storytelling, and informal interaction are overwhelmingly realized in CA, as seen in examples like "واحده قالت زوجيا إتاخر..." 'A woman said, 'My husband was late…', which rely on narrative style, exaggeration, and culturally embedded expressions. Such posts demonstrate that CA is the preferred medium for humor and affective engagement, particularly on Facebook, where extended narratives are more common. Another salient theme is *business* (and advertising) discourse, which appears across both platforms but shows different linguistic tendencies. On Facebook, CA is frequently used in promotional and advisory contexts, as in "صراحة تعبتهم حيل خلال هذا الشهر..." 'Honestly, I really exhausted them during this month…', where the speaker adopts an informal, persuasive tone to engage with audiences. On X, however, business and technology-related posts, such as "Samsung تضرب بقوة Apple…" *'Samsung is hitting Apple hard'*, tend to favor SA or a mixed register, reflecting a more formal and informational style aligned with broader audience reach. Knowledge-based discourse, i.e. science of language, is consistently expressed in SA, as in " التداولية هي دراسة اللغة..." *'pragmatics is the study of language…'*. This confirms the strong association between SA and formal, specialized knowledge domains across both platforms.

To recapitulate, the thematic analysis demonstrates that while both platforms host similar discourse domains, their distribution and linguistic realization differ significantly. X is characterized by a concentration of political and informational themes expressed predominantly in SA, whereas Facebook exhibits a broader thematic range, with strong representation of cultural, personal, and entertainment content realized primarily in CA. These patterns confirm that diglossic choice in ADD is not random but systematically linked to thematic context, reinforcing the notion of digital diglossia as a function of both platform affordances and discourse domains.

### 5.2. Digital analysis

To operationalize the proposed DDA approach, the dataset was systematically analyzed across 3 interrelated levels: micro (linguistic features), macro (discourse functions), and digital discourse (platform affordances). Table 7 exemplifies and summarizes all these aspects.

**Tabel 7: Digital features**

| Platform | Variety | Example | Micro features | Macro function | Digital features |
|---|---|---|---|---|---|
| X | SA | سيُكتب على أبواب غزة حكاية شعب انتصر 💪 🙏 🙏 🙏 | Passive; formal lexicon; SA orthography | Political/national narrative; authority; pan-Arab identity | Concise information-dense; Accommodating optimism; stance, pan-Arab identity |
| X | CA | تنام عشان تنسى همومك 🤣 🤣 | Dialectal marker "عشان"; simplified syntax | Personal reflection; affective stance | Narrative tone; engagement-oriented |
| Facebook | SA | التداولية هي دراسة اللغة في سياق استعمالها | Technical lexicon; definitional structure | Knowledge dissemination; educational discourse | Knowledge; low interactivity markers |
| Facebook | CA | صراااحة تعبتهم حيل خلال هذا الشهر 🤣 🤣 🤣 | Dialectal lexicon (حيل); spoken-like syntax | Personal experience; relatability; identity | Long-form narrative; audience engagement |
| X | SA | بايدن: سياسة نتنياهو خاطئة | Formal register; reporting structure | News reporting; institutional voice | Headline-like brevity |
| X | CA | صباح الرضى وكل الأماني الطيبة صباح الفل والياسمين❤️❤️ ❤️ | CS; coordination; CA lexicon | Culture; greeting; optimism | ADD accommodates everyday rituals |
| Facebook | CA | وضع الماء في الزعفرانيه يوم اكو و 10 ماكو 😭 😭 | Dialectical marker "اكو" | A cultural concept delivered by "الزعفرانيه" as an inherited value. | Accommodating heritage; ADD openness |
| Facebook | SA | أيها الاخوة لابد من الوحدة ولم الشمل لكي نتغلب على كل المشكلات التي تواجهنا، فالوحدة قوة 💪 | أيهاis a vocative marker; used for specificity and praise | Arab nationalism; Arab unity; | ADD call against in-divide; tomorrow hope |
| X | SA | بل سياتي يوم يجمع كل العرب 💪 💪 🙏 🙏 🙏 | بلa SA marker of more formality and solidarity | Arab nationalism; unity; Arab slogan | ADD does not abandon these markers but incorporates them |

| X | CA | ايوة لاااازم نشوفه اليوم قبل بكرة 💪 👍 | A dialectical marker "ايوة" for SA "لكن" 'but' | Informal use; closeness between the speaker and hearer | ADD openness; orthography; repetition of aleph اااا |
|---|---|---|---|---|---|

Table 7 presents a sample of DDA features that could be employed in the analysis of digital features. Our aim is to extend the thematic analysis to involve and operationalize DDA in our analysis. Thus, in what follows we attempt to undertake a systematic DDA by integrating micro-, macro-, and digital-level features. This multi-layered approach demonstrates how diglossic variation in SA vs. CA is not only linguistically structured but also shaped by platform affordances and communicative practices. For example, at the micro-level the data reveal consistent and systematic distinctions between SA and CA. SA usage is characterized by canonical morphosyntactic structures, formal lexical choices, and standardized orthography. This is evident in examples such as "سيُكتب على أبواب غزة حكاية شعب انتصر"*'history will write on gaza's doos a tale of people's victory*, which employs passive construction, formal lexicon, and standard spelling conventions. However, CA is marked by dialect-specific lexical items (e.g., *"اكو"، "حيل"، "عشان"*), simplified syntactic structures, and non-standard orthographic practices, including spacing variation and repetition, as in ""جااااالك تجارة اسكندرية"" and "ايوة لاااازم نشوفه اليوم قبل بكرة", repetition of و as in مبرووووووك, and repetition of ي as in الجو جميييل In fact, this repetition feature could largely be categorized as CA-oriented feature, because we almost did not find it in SA tweets/posts. Additional discourse markers such as "يبدو"، "بل"، "ترا", "طيب" are used. The former two are SA-oriented, and further signal pragmatic nuance and speaker stance. While in SA "يبدو" indicates uncertainty, " بل" is a cohesive device, which signals contrast and rhetorical emphasis. In addition, SA maintains morphosyntactic stability through well-formed nominal and verbal, nominal and clausal structures, contributing to clarity and formality. In CA "ترا " reflects advice, superiority of the speaker and his closeness to the audience, while "طيب" may point to inferiority of the speaker but also agreement, satisfaction, etc. Some of these linguistic devices are or come as a result of digital discourse, which we discuss later in this section. Anther linguistic feature that could be noted here is preciseness and a headline-like structure as in the tweet "بايدن: سياسة نتنياهو خاطئة".

At the macro-level, SA seems to be the pan-Arab device to express one's identity, stance, collective opinion, concern and commitment. The tweet "سيُكتب على أبواب غزة حكاية شعب انتصر" simply reflects this attitude of all Arabs. The SA variety does not only express formality, institutional factors, but also collective fear, grief, concern. It is a pan-Arab device due to the fact that some CA varieties may be unintelligible to each other (see also Shormani and Al Hussen 2024). On educational discourse, SA is also considered the medium of instruction for several and various knowledge disciplines across Arab world. These examples are associated with institutional, political discourse, indexing authority, formality, and pan-Arab identity. For instance, the tweet " *بايدن: سياسة نتنياهو خاطئة*" reflects a reporting structure typical of news discourse, or otherwise, a political discourse. Another aspect of SA concerns Arab nationalism. The tweet "بل سياتي يوم يجمع كل العرب!", for instance, expresses Arab nationalism, a dream of all Arab nations to unite; it is also an Arab slogan. CA, on the other hand, could be called a within-country device, i.e. to express such factors as identity, stance collective opinion, concern, but within the same country, community. For example, the post "وضع الماء في الزعفرانيه يوم اكو و 10 ماكو" expresses how day-to-day concerns are signaled on ADD, that Arab situation is rather pathetic in some countries, that people cannot find a water constant supplier, a substantial life ingredient. The term "الزعفرانيه" which is a cultural zone in Baghdad, refers to an ancient cultural inherited name survived from our predecessors. CA is also a means of interrelated concerns that express optimism, hope, though living in real misery. For example, the tweet "صباح الفل والياسمين صباح الرضية" could be viewed as a morning prayer, opening a new day of optimism, and a hope for a better tomorrow.

At the digital level, platform affordances play a decisive role in shaping discourse. X favors brevity, conciseness, and information density, resulting in a broadcast-oriented style that aligns with SA usage. Posts such as *"بايدن: سياسة نتنياهو خاطئة"* reinforce users' their informational and authoritative function. Digital communication has come up with several and various orthographical devices and writing techniques. For instance, digital practices such as orthographic play, abbreviation, emojis, and spacing variation, like 2 or more or 3 dots, function as meaningful pragmatic/semiotic resources. Examples include the use of abbreviated forms such as *"ج.م."* for *"جمعة مباركة"*, letter reduction ع = على, م = ما, for instance, and expressive repetition *"لاااازم"*, all of which enhance engagement and signal informality. Similarly, spacing expresses an unended message that may urge audience to question, guess and/or imagine what the speaker hides. It could also be a means of manipulation on the part of the speaker not to declare something (see also Omar and Ilyas 2018).

Another, perhaps the most important ADD feature digital age has come up with is the use of emojis. Emojis are digital pragmatic nonverbal cues used for enriching online communication, thus, compensating for the absence of facial expressions, gestures, and/or tone of voice used in face-to-face communication. They are used for a number of purposes including: i) signaling a difficult propositional attitude, ii) intensifying a propositional attitude coded verbally, iii) strengthening the illocutionary force of a speech act, vi) contradicting the explicit content of the utterance, and v) adding a feeling or emotion as decoration (Yus 2014: 51; Li and Yang 2018: 3, see also Yus 2025). In our case, to exemplify the use of emojis consider the emojis used in Table 5. For example, the 💪signals strength, well determination, which has been used in tweets like " بل سياتي يوم يجمع كل العرب" to emphasize the textual meaning, i.e. our strength is in our oneness/unity. The emoji 🤣expresses great laughter and fun as in "صرااااحة تعبتهم حيل خلال هذا الشهر" (cf. Aljasir 2020).

Our empirical findings suggest that the data undermine stability, challenging the classical notion of diglossic stability proposed by Ferguson (1959), demonstrating that digital communication actively disrupts the traditionally rigid functional division between H and L varieties. Rather than adhering to fixed, domain-specific boundaries, the data reveal a fluid and dynamic continuum on social media where SA and CA frequently overlap, blend, and cross functional contexts. Ultimately, the interactive affordances of digital platforms erode this historical equilibrium, replacing structural stability with flexible, hybrid linguistic practices (cf. also Bassiouney 2009; Khamis-Dakwar and Froud 2019). The quantitative results provide direct empirical validation for these qualitative thematic patterns and digital discourse features observed across the corpus. For example, bivariate associations and logistic regression models demonstrate that diglossic variation in Arabic social media is governed by an interplay between platform architecture and domain-specific formality norms. The Chi-square analyses confirm a robust baseline association between discourse domains and language choice across both platforms ($\chi^2 = 1883.03$, $p < .001$, $CV = .434$), with Facebook displaying a stronger overall dependence on domain ($CV = .500$) compared to X ($CV = .347$). This reflects broader sociolinguistic patterns: formal or institutionalized discourse domains, such as Politics and Science, strongly favor SA, whereas informal domains like Fun and Sports shift heavily towards CA. However, the main effect of platform ($\chi^2 = 262.16$, $p < .001$, $CV = .162$; $B = 0.2693$, $p = .0028$, $OR = 1.3090$) reveals that X maintains a consistently higher baseline probability of SA usage than Facebook. This baseline divergence aligns with X's public-facing, microblogging dynamic, which encourages formal register preservation, whereas Facebook's networked privacy and relational focus cultivate localized, informal CA interactions.

Beyond overall platform baselines, the interaction terms in the logistic regression model highlight how platform affordances actively reshape domain-specific language norms. While informal categories like Fun, Sports, and Culture show strong negative main effects for SA on Facebook (*ORs* < 0.21), these domain-level constraints are substantially mitigated on X. Specifically, the significant positive interaction terms for X in Culture ($OR = 2.6533$, $p < .001$), Fun ($OR = 6.3351$, $p < .001$), and Sports (OR = 26.7091, $p < .001$) indicate a marked shift towards SA when these topics are discussed on X relative to Facebook. Conversely, technical categories such as Science ($OR = 0.4088$, $p < .001$) and Technology ($OR = 0.7113$, $p = .0322$) exhibit negative interaction effects on X, suggesting a preference for localized or colloquial expressions in microblogged technical exchanges. Together, these findings indicate that diglossic choice is not determined solely by topic matter, but by a dynamic interaction where platform-specific affordances recalibrate traditional domain boundaries.

To recapitulate, these findings demonstrate that Arabic diglossia in online environments is best understood as digital diglossia: a multi-layered phenomenon in which linguistic variation is shaped not only by structural micro and functional macro factors but also by platform-specific affordances at the digital level. Additionally, representing the diglossia of ADD, it seems that DA has brought about a considerable change to Arabic diglossia. We notice several aspects: i) use of emojis/symbols to express (dis)like, agreement, emphasis, fear, hope, optimism. This is in line with Shormani and Alenezi's (2026) findings, ii) DA results in repetition of letters to show emphasis, ii) CA allows both X and Facebook users freedom, a freedom that is not allowed in SA restricted and rigid rules, iii) CA dialect coverage might be due to its reach to a wider audience, as SA is spoken by educated people and in formal situations, iv) SA remains dominant on both SMPs for its formality. These digital features illustrate how users exploit digital affordances to negotiate meaning, identity, stance and interaction, v), digital diglossia maintains traditional communication features and levels, and most importantly, vi), the study emphasizes that not only can digital diglossia preserves traditional communication, it can also "accommodate" new (digital) features, levels and styles DA has brought about.

However, this study involves some limitations: i) CA is treated in this study as a broad category encompassing colloquial varieties across the Arab world, but not as a set of regionally specified dialects. Given the considerable linguistic variation among Arabic dialects and their regional and subregional features, this general classification limits the extent to which the findings can account for dialect-specific patterns, ii) restricting the data sources to X and Facebook may not fully capture the diversity of ADD. Broader research could thus extend the analysis to other SMPs such as Telegram, and examine whether similar patterns occur across different digital environments, iii) we did not examine diglossic variation systematically across different genres. Although the corpus includes 7 discourse categories, genre-specific variation in the use of SA and CA was not investigated in detail, and iv) we did not involve or compare the countries or dialect regions of CA. Further research could incorporate country- or region-level metadata to examine whether the distribution and use of SA and CA vary across different communities and dialect regions. Future research could also address how Arabic diglossia varies across different genres and communicative contexts, providing a more fine-grained account of digital Arabic variation, and we leave these for future research.

## 6. Conclusions and implications

Thus, the collected posts/tweets reveal clear differences in topic, focus, style, and diglossic use across X and Facebook. The quantitative analysis shows distinct patterns in the use of SA and CA across the two social media platforms. On X, SA dominates, with 3033 tweets (60.7%) compared with 1967 CA tweets (39.3%). On Facebook, however, CA dominates, accounting for 2770 posts (55.4%), while SA accounts for 2230 posts (44.6%). Across the complete dataset of 10000 tweets/posts, SA accounts for 5263 cases (52.6%), compared with 4737 cases (47.4%) for CA. Thus, SA remains slightly more frequent overall, although the difference between the two varieties is relatively small. The binary logistic regression further clarifies these platform differences. With Facebook and Politics as the reference categories, the main platform effect was significant ($B$ = 0.2693, $p$ = .0028, $OR$ = 1.3090, 95% CI [1.0974, 1.5615]), indicating higher odds of SA use on X than on Facebook within the Politics reference category. More importantly, significant Platform × Discourse Category interactions demonstrate that the relationship between discourse category and diglossic choice varies across platforms. Significant interactions were found for Culture ($OR$ = 2.6533, $p$ < .001), Fun ($OR$ = 6.3351, $p$ < .001), Sports ($OR$ = 26.7091, $p$ < .001), Science ($OR$ = 0.4088, $p$ < .001), and Technology ($OR$ = 0.7113, $p$ = .0322). The Business interaction was not significant ($OR$ = 1.1922, p = .2412). These results indicate that platform and discourse category jointly structure patterns of diglossic choice, and not suggesting a uniform platform effect across all topics.

The distributions are consistent with this interaction pattern. On X, SA is particularly prominent in Politics (75.4%), Technology (67.0%), Culture (62.9%), Sports (59.6%), and Science (56.4%), whereas CA is more frequent in Business (53.3%) and Fun (85.9%). On Facebook, SA is most frequent in Science (70.7%), Technology (68.6%), and Politics (70.1%), while CA predominates in Culture (67.2%), Business (64.1%), Fun (98.1%), and Sports (95.9%). These distributions show that the relationship between diglossic choice and discourse category is not identical across the two platforms. The findings thus suggest that digital diglossia is associated with the intersection of platform and communicative domain, but not with a uniform platform effect. X shows a higher overall proportion of SA, whereas Facebook shows a higher overall proportion of CA; however, the regression results demonstrate that this difference varies according to discourse category. The particularly large interaction effects for Fun and Sports, for instance, indicate substantial platform differences in the association between these domains and SA/CA choice. However, the negative interactions for Science and Technology indicate a different platform pattern for these domains.

These patterns may partly reflect differences in communicative context. X may provide greater visibility for public, institutional, or professional discourse, whereas Facebook may accommodate more personal and socially embedded interaction. Such interpretations, however, should be treated as possible contextual explanations rather than as causal effects established by the present statistical analysis. The results are consistent with the broader sociolinguistic observation that Arabic users may favor SA in more formal or public contexts while using CA more extensively in informal and socially oriented communication (Ferguson 1959). Recent research on social-media diglossia similarly reports greater use of SA in areas such as politics, religion, and public discourse, while CA is more prevalent in informal areas such as sports, fun, and personal matters (Yuldasheva and Sidikova 2025). When the dataset is considered as a whole, the relatively close proportions of SA (52.6%) and CA (47.4%) indicate that neither variety overwhelmingly dominates digital written Arabic across the two platforms. At the same time, the platform-specific distributions demonstrate the continuing relevance of traditional diglossic differentiation in digital communication. SA remains strongly represented in domains such as politics, science, and technology, while CA is especially prevalent in more conversationally oriented domains such as fun, culture, business, and sports on Facebook. These patterns suggest that digital media provide expanded spaces for the written use of CA while SA continues to occupy an important role across platforms and discourse domains.

**Competing interests**
The authors declare no competing interests.

**Data Availability**
The datasets generated and analyzed during the current study are provided as supplementary material.